# Feature Selection By KDDA For SVM-Based MultiView Face Recognition

Seyyed Majid Valiollahzadeh, Abolghasem Sayadiyan, Mohammad Nazari

*Electrical Engineering Department, Amirkabir University of Technology,
Tehran, Iran, 15914*

valiollahzadeh@yahoo.com

eea35@aut.ac.ir

mohnazari@aut.ac.ir

**Abstract:** Applications such as Face Recognition (FR) that deal with high-dimensional data need a mapping technique that introduces representation of low-dimensional features with enhanced discriminatory power and a proper classifier, able to classify those complex features .Most of traditional Linear Discriminant Analysis (LDA) suffer from the disadvantage that their optimality criteria are not directly related to the classification ability of the obtained feature representation. Moreover, their classification accuracy is affected by the "small sample size" (SSS) problem which is often encountered in FR tasks. In this short paper, we combine nonlinear kernel based mapping of data called KDDA with Support Vector machine (SVM) classifier to deal with both of the shortcomings in an efficient and cost effective manner. The proposed here method is compared, in terms of classification accuracy, to other commonly used FR methods on UMIST face database. Results indicate that the performance of the proposed method is overall superior to those of traditional FR approaches, such as the Eigenfaces, Fisherfaces, and D-LDA methods and traditional linear classifiers.

**Keywords:** Face Recognition, Kernel Direct Discriminant Analysis (KDDA), small sample size problem (SSS), Support Vector Machine (SVM).

## INTRODUCTION

Selecting appropriate features to represent faces and proper classification of these features are two central issues to face recognition (FR) systems. For feature selection, successful solutions seem to be appearance-based approaches, (see [3], [2] for a survey), which directly operate on images or appearances of face objects and process the images as two-dimensional (2-D) holistic patterns, to avoid difficulties associated with Three-dimensional (3-D) modelling, and shape or landmark detection [2]. For the purpose of data reduction and feature extraction in the appearance-based approaches, Principle component analysis (PCA) and linear discriminant analysis (LDA) are introduced as two powerful tools. Eigenfaces [4] and Fisherfaces [5] built on the two techniques, respectively, are two state-of-the-art FR methods, proved to be very successful. It is generally believed that, LDA based algorithms outperform PCA based ones in solving problems of pattern classification, since the former optimizes the low-dimensional representation of the objects with focus on the most discriminant feature extraction while the latter achieves simply object reconstruction. However, many LDA based algorithms suffer from the so-called "small sample size problem" (SSS) which exists in high-dimensional pattern recognition tasks where the number of available samples is smaller than the dimensionality of the samples. The traditional solution to the SSS problem is to utilize PCA concepts in conjunction with LDA (PCA+LDA) as it was done for example in Fisherfaces [11]. Recently, more effective solutions, called Direct LDA (D-LDA) methods, have been presented [12], [13]. Although successful in many cases, linear methods fail to deliver good performance when face patterns are subject to large variations in viewpoints, which results in a highly non-convex and complex distribution. The limited success of these methods should be attributed to their linear nature [14]. Kernel discriminant analysis algorithm, (KDDA) generalizes the strengths of the recently presented D-LDA [1] and the kernel techniques while at the same time overcomes many of their shortcomings and limitations.

In this work, we first nonlinearly map the original input space to an implicit high-dimensional feature space, where the distribution of face patterns is hoped to be linearized and simplified. Then, KDDA method is introduced to effectively solve the SSS problem and derive a set of optimal discriminant basis vectors in the feature space. And then SVM approach is used for classification.

The rest of the paper is organized as follows. In Section tow, we start the analysis by briefly reviewing KDDA method. Following that in section three, SVM is introduced and analyzed as a powerful classifier. In Section four, a set of experiments are presented to demonstrate the effectiveness of the KDDA algorithm together with SVM classifier on highly nonlinear, highly complex face pattern distributions. The proposed method is compared, in terms of the classification error rate performance, to KPCA (kernel based PCA), GDA (Generalized Discriminant Analysis) and KDDA algorithm with nearest neighbour classifier on the multi-view UMIST face database. Conclusions are summarized in Section five.

## 2 Kernel Direct Discrimi-nant Analysis (KDDA)

### 2.1 Linear Discriminant Analysis

In the statistical pattern recognition tasks, the problem of feature extraction can be stated as follows: Assume that we have a training set, $\{Z_i\}_{i=1}^{L}$ is available. Each image is defined as a vector of length $N (= I_w \times I_h)$, i.e. $Z_i \in \Re^N$ where $I_w \times I_h$ is the face image size and $\Re^N$ denotes a N-dimensional real space [1].

It is further assumed that each image belongs to one of C classes $\{Z_i\}_{i=1}^{C}$. The objective is to find a transformation $\varphi$, based on optimization of certain separability criteria, which produces a mapping, with $y_i \in \Re^N$ that leads to an enhanced separability of different face objects.

Let $S_{BTW}$ and $S_{WTH}$ be the between- and within-class scatter matrices in the feature space $\mathbb{F}$ respectively, expressed as follows:

$$S_{BTW} = \frac{1}{L}\sum_{i=1}^{C} C_i (\overline{\phi}_i - \overline{\phi})(\overline{\phi}_i - \overline{\phi})^T \quad (1)$$

$$S_{WTH} = \frac{1}{L}\sum_{i=1}^{C}\sum_{j=1}^{C_i} (\overline{\phi}_{ij} - \overline{\phi}_i)(\overline{\phi}_{ij} - \overline{\phi}_i)^T \quad (2)$$

Where $\phi_{ij} = \phi(Z_{ij})$, $\overline{\phi}_i$ is the mean of class $Z_i$ and $\overline{\phi}$ is the average of the ensemble.

$$\overline{\phi}_i = \frac{1}{C_i}\sum_{j=1}^{C_i} \phi(Z_{ij}) \quad (3)$$

$$\overline{\phi} = \frac{1}{L}\sum_{i=1}^{C}\sum_{j=1}^{C_i} \phi(Z_{ij}) \quad (4)$$

The maximization can be achieved by solving the following eigenvalue problem:

$$\Phi = \arg\max_{\Phi} \frac{|\Phi^T S_{BTW} \Phi|}{|\Phi^T S_{WTH} \Phi|} \quad (5)$$

The feature space F could be considered as a "linearization space" [6], however, its dimensionality could be arbitrarily large, and possibly infinite. Solving this problem lead us to LDA[1].

Assuming that is $S_{WTH}$ nonsingular and $\Phi$ the basis vectors correspond to the M first eigenvectors with the largest eigenvalues of the discriminant criterion:

$$J = tr(S_{WTH}^{-1} S_{BtW} \Phi) \quad (6)$$

The M-dimensional representation is then obtained by projecting the original face images onto the subspace spanned by the eigenvectors.

### 2.2 Kernel Direct Discriminant Analysis (KDDA)

The maximization process in (3) is not directly linked to the classification error which is the criterion of performance used to measure the success of the FR procedure. Modified versions of the method, such as the Direct LDA (D-LDA) approach, use a weighting function in the input space, to penalize those classes that are close and can potentially lead to misclassifications in the output space.

Most LDA based algorithms including Fisherfaces [7] and D-LDA [9] utilize the conventional Fisher's criterion denoted by (3).

The introduction of the kernel function allows us to avoid the explicit evaluation of the mapping. Any function satisfying Mercer's condition can be used as a kernel, and typical kernel functions include polynomial function, radial basis function (RBF) and multi-layer perceptrons [10].

$$\Phi = \arg\max_{\Phi} \frac{|\Phi^T S_{BTW} \Phi|}{|(\Phi^T S_{BTW} \Phi) + (\Phi^T S_{WTH} \Phi)|} \quad (7)$$

The KDDA method implements an improved D-LDA in a high-dimensional feature space using a kernel approach.

KDDA introduces a nonlinear mapping from the input space to an implicit high dimensional feature space, where the nonlinear and complex distribution of patterns in the input space is "linearized" and "simplified" so that conventional LDA can be applied and it effectively solves the small sample size (SSS) problem in the high-dimensional feature space by employing an improved D-LDA algorithm.

Unlike the original D-LDA method of [10] zero eigenvalues of the within-class scatter matrix are never used as divisors in the improved one. In this way, the optimal discriminant features can be exactly extracted from both of inside and outside of $S_{WTH}$'s null space.

In GDA, to remove the null space of $S_{WTH}$, it is required to compute the pseudo inverse of the kernel matrix K, which could be extremely ill-conditioned when certain kernels or kernel parameters are used. Pseudo inversion is based on inversion of the nonzero eigenvalues.

## 3 SVM Based Approach for Classification

The principle of Support Vector Machine (SVM) relies on a linear separation in a high dimension feature space where the data have been previously mapped, in order to take into account the eventual non-linearities of the problem.

### 3.1 Support Vector Machines (SVM)

If we assume that, the training set $X = (x_i)_{i=1}^{l} \subset \mathrm{R}^R$ where $l$ is the number of training vectors, R stands for the real line and R is the number of modalities, is labelled with two class targets $Y = (y_i)_{i=1}^{l}$, where:

$$y_i \in \{-1,+1\} \quad \Phi : \mathrm{R}^R \to F \qquad (8)$$

Maps the data into a feature space F. Vapnik has proved that maximizing the minimum distance in space F between $\Phi(X)$ and the separating hyper plane $H(w,b)$ is a good means of reducing the generalization risk. Where:

$$H(w,b) = \{f \in F \mid <w,f>_F + b = 0\},$$
$$(<> \text{ is inner product}) \qquad (9)$$

Vapnik also proved that the optimal hyper plane can be obtained solving the convex quadratic programming (QP) problem:

$$\text{Minimize} \quad \frac{1}{2}\|w\|^2 + c\sum_{i=1}^{l} \xi_i$$
$$\text{with} \quad y_i(<w,\Phi(X)> + b) \geq 1 - \xi_i \quad i=1,...,l \qquad (10)$$

Where constant C and slack variables x are introduced to take into account the eventual non-separability of $\Phi(X)$ into F.

In practice this criterion is softened to the minimization of a cost factor involving both the complexity of the classifier and the degree to which marginal points are misclassified, and the tradeoff between these factors is managed through a margin of error parameter (usually designated C) which is tuned through cross-validation procedures. Although the SVM is based upon a linear discriminator, it is not restricted to making linear hypotheses. Non-linear decisions are made possible by a non-linear mapping of the data to a higher dimensional space. The phenomenon is analogous to folding a flat sheet of paper into any three-dimensional shape and then cutting it into two halves, the resultant non-linear boundary in the two-dimensional space is revealed by unfolding the pieces.

The SVM's non-parametric mathematical formulation allows these transformations to be applied efficiently and implicitly: the SVM's objective is a function of the dot product between pairs of vectors; the substitution of the original dot products with those computed in another space eliminates the need to transform the original data points explicitly to the higher space. The computation of dot products between vectors without explicitly mapping to another space is performed by a kernel function.

The nonlinear projection of the data is performed by this kernel functions. There are several common kernel functions that are used such as the linear, polynomial kernel $(K(x, y) = (<x, y>_{R^R} + 1)^d$ and the sigmoidal kernel $(K(x, y) = \tanh(<x, y>_{R^R} + a))$, where x and y are feature vectors in the input space.

The other popular kernel is the Gaussian (or "radial basis function") kernel, defined as:

$$K(x, y) = \exp\left(\frac{-|x-y|^2}{(2\sigma^2)}\right) \qquad (11)$$

Where $\sigma$ is a scale parameter, and x and y are feature-vectors in the input space. The Gaussian kernel has two hyper parameters to control performance C and the scale parameter $\sigma$. In this paper we used radial basis function (RBF).

### 3.2 Multi-class SVM

The standard Support Vector Machines (SVM) is designed for dichotomic classification problem (two classes, called also binary classification).

Several different schemes can be applied to the basic SVM algorithm to handle the K-class pattern classification problem. These schemes will be discussed in this section. The K-class pattern classification problem is posted as follow:

- Given $l$ i.i.d. sample: $(x_1, y_1), ..., (x_l, y_l)$ where $x_i$, for $i=1,...,l$ is a feature vector of length d and $y_i = \{1,...,k\}$ is the class label for data point $x_i$.

- Find a classifier with the decision function, $f(x)$ such that $y = f(x)$ where y is the class label for $x$.

The multi-class classification problem is commonly solved by decomposition to several binary problems for which the standard SVM can be used.

For solving the multi-class problem are as listed below:

- Using K one-to-rest classifiers (one-against-all)

- Using $k(k-1)/2$ pair wise classifiers
- Extending the formulation of SVM to support the k-class problem.

### 3.2.1. Combination of one-to-rest classifiers

This scheme is the simplest, and it does give reasonable results. K classifiers will be constructed, one for each class. The K-th classifier will be trained to classify the training data of class k against all other training data. The decision function for each of the classifier will be combined to give the final classification decision on the K-class classification problem. In this case the classification problem to k classes is decomposed to k dichotomy decisions $f_m(x)$, $m \in K = 1,...,k$ where the rule $f_m(x)$ separates training data of the m-th class from the other training patterns. The classification of a pattern x is performed according to maximal value of functions $f_m(x)$, $m \in K$, $K = 1,...,k$ i.e. the label of $x$ is computed as:

$$f(x) = \arg(\max_{m \in k}(f_m(x))) \qquad (12)$$

### 3.2.2. Pair wise Coupling classifiers

The schemes require a binary classifier for each possible pair of classes. The decision function of the SVM classifier for $y_1$-to-$y_2$ and $y_2$-to-$y_1$ has reflectional symmetry in the zero planes. Hence only one of these pairs of classifier is needed. The total number of classifiers for a K-class problem will then be $k(k-1)/2$. The training data for each classifier is a subset of the available training data, and it will only contain the data for the two involved classes. The data will be reliable accordingly, i.e. one will be labeled as +1 while the other as -1. These classifiers will now be combined with some voting scheme to give the final classification results. The voting schemes need the pair wise probability, i.e. the probability of x belong to class i given that it can be only belong to class i or j.

The output value of the decision function of an SVM is not an estimate of the p.d.f. of a class or the pair wise probability. One way to estimate the required information from the output of the SVM decision function is proposed by (Hastie and Tibshirani, 1996) The Gaussian p.d.f. of a particular class is estimated from the output values of the decision function, $f(x)$, for all x in that class. The centroid and radius of the Gaussian is the mean and standard deviation of $f(x)$ respectively.

## 4 EXPERIMENTS AND RESULTS

### 4.1 Database

In our work, we used a popular face databases (The UMIST [13]), for demonstrating the effectiveness of our combined KDDA and SVM proposed method. It is compared with KPCA, GDA and KDDA algorithm with nearest neighbor classifier.

We use a radial basis function (RBF) kernel function:

$$K(x, y) = \exp(\frac{-|x-y|^2}{(2\sigma^2)}) \qquad (13)$$

Where $\sigma$ is a scale parameter, and x and y are feature-vectors in the input space. The RBF function is selected for the proposed SVM method and KDDA in the experiments. The selection of scale parameter $\sigma$ is empirical.

In addition, in the experiments the training set is selected randomly each time, so there exists some fluctuation among the results. In order to reduce the fluctuation, we do each experiment more than 10 times and use the average of them.

### 4.2 UMIST Database

The UMIST repository is a multi-view database, consisting of 575 images of 20 people, each covering a wide range of poses from profile to frontal views. Figure 1 depicts some samples contained in the two databases, where each image is scaled into (112 92), resulting in an input dimensionality of N = 10304.

For the face recognition experiments, in UMIST database is randomly partitioned into a training set and a test set with no overlap between the two set. We used ten images per person randomly chosen for training, and the other ten for testing. Thus, training set of 200 images and the remaining 375 images are used to form the test set.

It is worthy to mention here that both experimental setups introduce SSS conditions since the number of training samples are in both cases much smaller than the dimensionality of the input space [1].

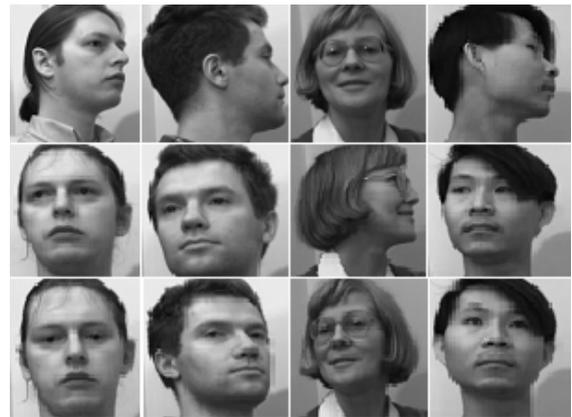

Figure 1: Some sample images of four persons randomly chosen from the UMIST database.

On this database, we test the methods with different training samples and testing samples corresponding the training number k=2, 3, 4, 5,6,7,8 of each subject. Each time randomly select k samples from each subject to train and the other $10-K$ to test. The experimental results are given in the table 1.

Table 1. Recognition rate (%) on the UMIST database.

| K | Our method (KDDA+SVM) | KDDA +NN * | KPCA | GDA |
|---|---|---|---|---|
| 2 | 81.8 | 81.9 | 75.5 | 71.5 |
| 3 | 83.5 | 83.4 | 76.2 | 72.8 |
| 4 | 87.3 | 85.4 | 77.1 | 74.5 |
| 5 | 90.4 | 87.9 | 79.8 | 75.1 |
| 6 | 94.1 | 89.1 | 83.4 | 79.0 |
| 7 | 96.0 | 93.9 | 87.1 | 82.1 |
| 10 | 96.5 | 95.2 | 89.1 | 83.0 |

* Nearest Neighbour

Figure 2 depicts the first two most discriminant features extracted by utilizing KDDA respectively and we show the decision boundary for first 6 classes for training data in Combination of one-to-rest classifier SVM.

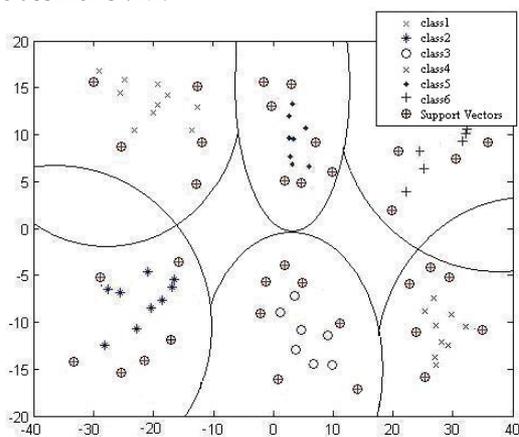

Figure 2: The decision boundary for first 6 classes for training data (Combination of one-to-rest classifier SVM)

The only kernel parameter for RBF is the scale value $\sigma^2$ for SVM classifier. Figure.4 shows the error rates as functions of $\sigma^2$, when the optimal number of feature vectors (M is optimum) is used.

As such, the average error rates of our method with RBF kernel are shown in Figure 5. It shows the error rates as functions of M within the range from 2 to 19 ($\sigma^2$ is optimum).

## 5 Discussions and Conclusions

A new FR method has been introduced in this paper. The proposed method combines kernel-based methodologies with discriminant analysis techniques and SVM classifier. The kernel function is utilized to map the original face patterns to a high-dimensional feature space, where the highly non-convex and complex distribution of face patterns is simplified, so that linear discriminant techniques can be used for feature extraction.

The small sample size problem caused by high dimensionality of mapped patterns is addressed by a kernel-based D-LDA technique (KDDA) which exactly finds the optimal discriminant subspace of the feature space without any loss of significant discriminant information.

Then feature space will be fed to SVM classifier. Experimental results indicate that the performance of the KDDA algorithm together with SVM is overall superior to those obtained by the KPCA or GDA approaches. In conclusion, the KDDA mapping and SVM classifier is a general pattern recognition method for nonlinearly feature extraction from high-dimensional input patterns without suffering from the SSS problem.

We expect that in addition to face recognition, KDDA will provide excellent performance in applications where classification tasks are routinely performed, such as content-based image indexing and retrieval, video and audio classification.

## Acknowledgements

The authors would like to acknowledge the Iran Telecommunication Research Center (ITRC) for financially supporting this work.

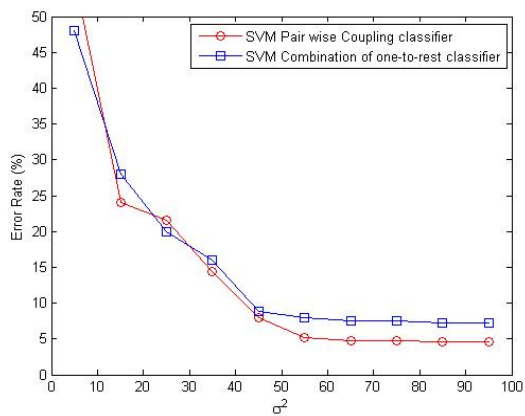

Figure 3: error rates as functions $\sigma^2$ of SVM. ($\sigma^2_{KDDA} = 5 \times 10^6$ [1])

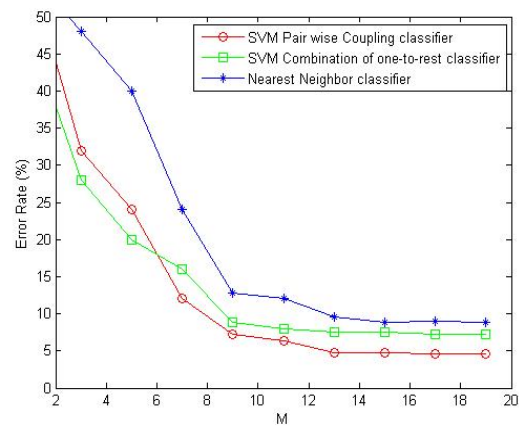

Figure 4: Comparison of error rates based on RBF kernel function.

We would also like to thank Dr. Daniel Graham and Dr. Nigel Allinson for providing the UMIST face database.

## References

[1] J. Lu, K. N. Plataniotis, A. N. Venetsanopoulos, "Face Recognition Using LDA-Based Algorithms" IEEE Trans. ON Neural Networks, vol. 14, no. 1, Jan.2003.

[2] M.Turk, "A random walk through eigenspace," *IEICE Trans. Inform.Syst.*, vol. E84-D, pp. 1586–1695, Dec. 2001.

[3] R. Chellappa, C. L.Wilson, and S. Sirohey, "Human and machine recognition of faces: A survey," *Proc. IEEE*, vol. 83, pp. 705–740, May 1995.

[4] M. Turk and A. P. Pentland, "Eigenfaces for recognition," *J. Cognitive Neurosci.*, vol. 3, no. 1, pp. 71–86, 1991.

[5] P. N. Belhumeur, J. P. Hespanha, and D. J. Kriegman, "Eigenfaces vs. Fisherfaces: Recognition using class specific linear projection," *IEEE Trans. Pattern Anal. Machine Intell.*, vol. 19, pp. 711–720, May 1997.

[6] M. A. Aizerman, E. M. Braverman, and L. I. Rozonoer, "Theoretical foundations of the potential function method in pattern recognition learning", Automation and Remote Control, vol. 25, pp. 821–837, 1964.

[7] L.-F.Chen, H.-Y. Mark Liao, M-.T. Ko, J.-C. Lin, and G.-J. Yu, "A new LDA-based face recognition system which can solve the small sample size problem," Pattern Recognition, vol. 33, pp. 1713–1726, 2000.

[9] H. Yu and J. Yang, "A direct LDA algorithm for high-dimensional data with application to face recognition," *Pattern Recognition*, vol. 34, pp. 2067–2070, 2001.

[10] V. N. Vapnik, "The Nature of Statistical Learning Theory", Springer-Verlag, New York, 1995.

[11] D. B. Graham and N. M. Allinson, "Characterizing virtual eigensignatures for general purpose face recognition," in Face Recognition: From Theory to Applications, H. Wechsler, P. J. Phillips, V. Bruce, F. Fogelman- Soulie, and T. S. Huang, Eds., 1998, vol. 163, NATO ASI Series F, Computer and Systems Sciences, pp. 446–456.

[12] D. L. Swets and J. Weng, "Using discriminant eigenfeatures for image retrieval," *IEEE Trans. Pattern Anal. Machine Intell.*, vol. 18, pp. 831–836, Aug. 1996.

[13] Q. Liu, R. Huang, H. Lu, S. Ma, "Kernel-Based Optimized Feature Vectors Selection and Discriminant Analysis for Face Recognition" 2002 IEEE

[14] C. Liu and H.Wechsler, "Evolutionary pursuit and its application to face recognition," *IEEE Trans. Pattern Anal. Machine Intell.*, vol. 22, pp. 570–582, June 2000.

[15] K. Liu, Y. Q. Cheng, J. Y. Yang, and X. Liu, "An efficient algorithm for Foley–Sammon optimal set of discriminant vectors by algebraic method," *Int. J. Pattern Recog. Artificial Intell.*, vol. 6, pp. 817–829, 1992.

[16] 0. Duda, R., E. Han. P., and *G*. Stork, D. *Parrern Recognirion.* John Wiley & Sons, 2000.

[17] L. Mangasarian. 0 . and R. Musicant. D. Successive over relaxation for support vector machines, *IEEE Transacrions on Neural Nerworks,* l0(5), 1999.